%% file: main.tex
\documentclass[10pt,twocolumn,letterpaper]{article}
\usepackage[T1]{fontenc}

\usepackage{booktabs}
\usepackage{tabularx}
\usepackage{verbatim}
\usepackage{float}
\usepackage{cuted}
\usepackage{caption}
\usepackage{wrapfig} 

\usepackage[table,dvipsnames]{xcolor}
\usepackage{graphicx}
\usepackage{multirow}

\usepackage{float}
\usepackage{algorithm}
\usepackage{algpseudocode}
\algrenewcommand\algorithmicrequire{\textbf{Input:}}
\algrenewcommand\algorithmicensure{\textbf{Output:}}
\usepackage{xcolor}
\usepackage{tabularx}
\usepackage{array}
\usepackage{subcaption}
\newcommand{\g}[1]{\textcolor{gray}{#1}}
\definecolor{wrongred}{RGB}{190,45,45}
\definecolor{correctgreen}{RGB}{35,135,70}

\usepackage[pagenumbers]{cvpr} 


\AtBeginDocument{%
  \appto\normalsize{%
    \setlength{\abovedisplayskip}{8pt plus 2pt minus 3pt}%
    \setlength{\belowdisplayskip}{8pt plus 2pt minus 3pt}%
    \setlength{\abovedisplayshortskip}{0pt plus 2pt}%
    \setlength{\belowdisplayshortskip}{4pt plus 2pt minus 2pt}%
  }%
  \normalsize
}

\definecolor{cvprblue}{rgb}{0.21,0.49,0.74}
\usepackage[pagebackref,breaklinks,colorlinks,allcolors=cvprblue]{hyperref}
\definecolor{referenceurl}{HTML}{EC008C}
\hypersetup{urlcolor=referenceurl}
\let\ViRAOriginalTTDefault\ttdefault
\usepackage[varl,noupquote]{inconsolata}
\let\ttdefault\ViRAOriginalTTDefault

\def\confName{CVPR}
\def\confYear{2027}

\title{ViRA: Agentic Recovery for Robust Video Understanding under Corruption}

\author{Bryan Gao$^{1*}$ \quad Ce Zhang$^{2*}$ \quad Bojiang Jin$^{3}$ \quad Weiguo Feng$^{3}$ \\[3pt]
$^{1}$University of Michigan \quad $^{2}$Carnegie Mellon University \quad $^{3}$TikTok Inc. \quad {$^{*}$Equal contribution} \\[3pt]
Project page: \url{https://bg3495.github.io/vira/}
}

\begin{document}

\maketitle

\input{sec/0_abstract}
\input{sec/1_intro}
\input{sec/2_related}
\input{sec/3_method}
\input{sec/4_experiments}
\input{sec/5_conclusion}


{
    \small
    \bibliographystyle{ieeenat_fullname}
    \bibliography{main}
}

\end{document}

%% file: sec/0_abstract.tex
\begin{strip}
\vspace{-48pt}
\centering
\includegraphics[width=\textwidth]{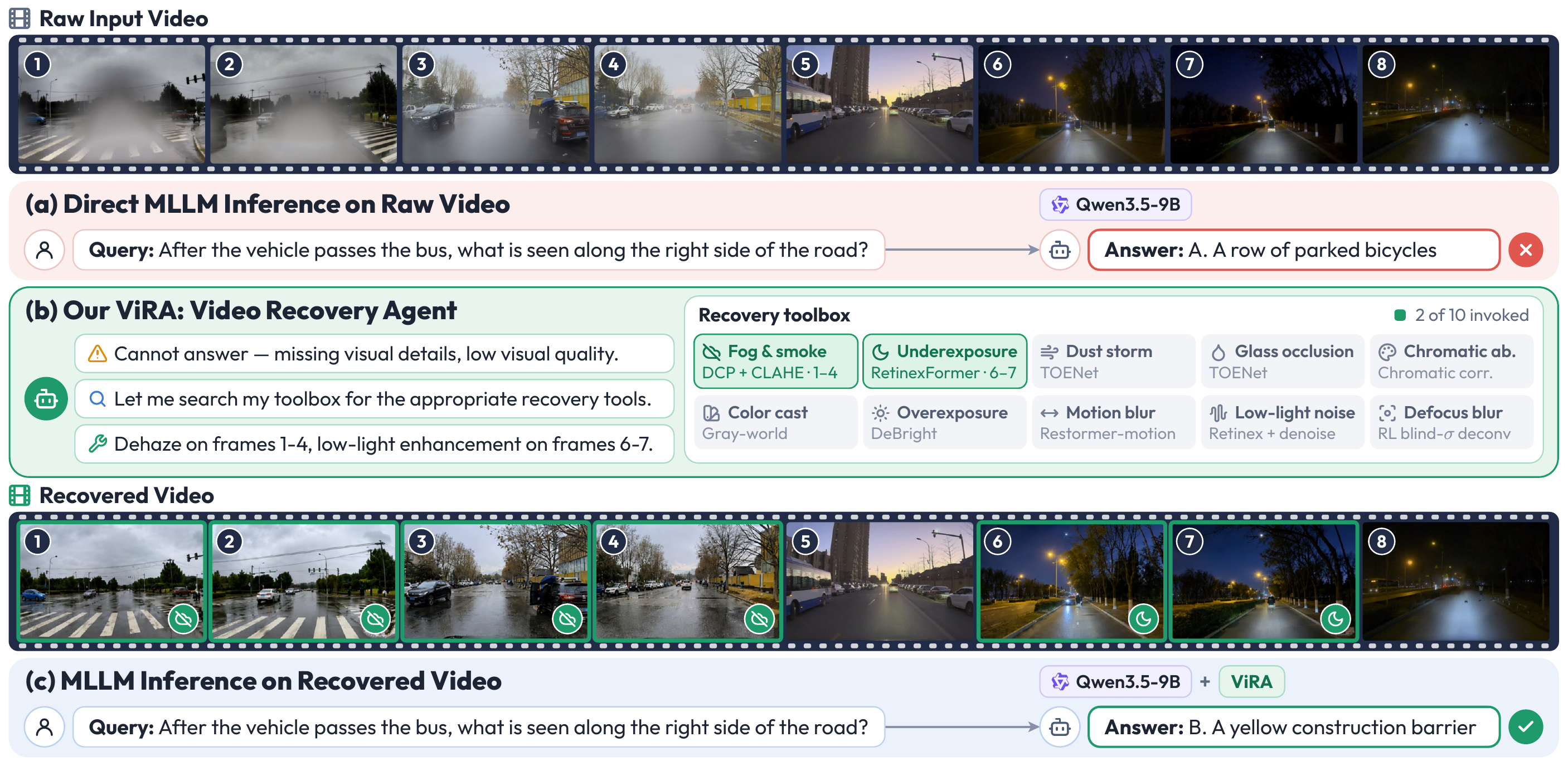}\par
\captionof{figure}{\textbf{ViRA recovers the visual evidence needed to answer a question by applying restoration tools to selected frames.} \textbf{(a)} Asked what appears on the right side of the road after the vehicle passes a bus, Qwen3.5-9B incorrectly reports a row of parked bicycles. \textbf{(b)} ViRA identifies two intervals where poor visibility obscures the roadside. It selects two of the ten tools in its toolbox, applying dehazing to frames 1--4 and low-light enhancement to frames 6--7. The other frames pass through unchanged, preserving the surrounding traffic scene and the order of events as the vehicle passes the bus. \textbf{(c)} Given the recovered video and the same question, Qwen3.5-9B correctly identifies the yellow construction barrier on the right side of the road.}
\label{fig:teaser}
\par
\end{strip}

\vspace*{-18pt}
\begin{abstract}

\enlargethispage{-18.18pt}\vspace{-6pt}
Video understanding models have achieved strong performance on clean, high-quality inputs, yet their robustness to real-world video degradations remains underexplored. In practice, videos are frequently affected by blur, noise, low illumination, haze, occlusion, and other spatially or temporally varying degradations that can obscure critical visual evidence and substantially impair downstream reasoning. Moreover, such degradations are often heterogeneous and localized, making uniform restoration both inefficient and potentially unnecessary. We introduce \textbf{ViRA}, a \textbf{Vi}deo \textbf{R}ecovery \textbf{A}gent that integrates visual recovery directly into the reasoning process. Rather than blindly enhancing an entire degraded video, ViRA follows a diagnose--restore--reason paradigm that jointly determines whether recovery is necessary, which restoration tools to invoke, and where in the video each tool should be applied. Conditioned on both the video and the downstream query, ViRA selectively recovers only the temporal regions whose degradation obscures task-relevant evidence, and then reasons over the recovered observation. Across extensive experiments on Video-MME, STAR, and NExT-QA, diverse degradation types, and MLLM backbones, ViRA improves accuracy under corruption by 3.0 points on average and by up to 9.5 points over direct inference. Importantly, these robustness gains do not come at the expense of clean-input performance: ViRA preserves and can even slightly improve accuracy on uncorrupted videos. \looseness=-1

\end{abstract}

%% file: sec/1_intro.tex
\section{Introduction}
\label{sec:intro}
Video understanding has become a central capability of modern multi-modal large language models (MLLMs), enabling applications ranging from long-form video question answering~\cite{zhang2026lens,he2024malmm} to embodied perception~\cite{di2024grounded} and autonomous driving~\cite{xu2025drivegpt4v2,lin2025jarvis}. Recent MLLMs~\cite{li2025llavaonevision,bai2025qwen2.5,wang2025internvl3} have demonstrated impressive reasoning ability on standard benchmarks, but most evaluations assume clean and high-quality visual inputs~\cite{fu2025videomme,fang2024mmbench}. This assumption rarely holds in practice: real-world videos are frequently affected by low illumination, motion and defocus blur, sensor noise, compression artifacts, adverse weather, and temporal disruptions~\cite{safonov2025ntireugc,li2025diffvsr,awad2026awdb}. Such degradations can obscure precisely the visual evidence needed for downstream reasoning, causing models that perform reliably on clean videos to fail under corrupted observations. As Video MLLMs are increasingly deployed in open-world environments, robustness to imperfect visual inputs is therefore becoming a fundamental requirement for reliable video understanding~\cite{lin2025jarvis,park2025resilient}.

Fortunately, a mature line of visual restoration research has shown that corrupted observations can often be effectively recovered using specialized methods, ranging from general restoration models such as Restormer~\cite{zamir2022restormer} to dedicated approaches for low-light enhancement~\cite{cai2023retinexformer} and adverse-weather restoration~\cite{gao2023toenet}. Similar progress has been made in video restoration, where temporal information across frames provides additional cues for recovering degraded content~\cite{chan2021basicvsr,liang2022rvrt,zhao2024avernet}. This opens up a natural opportunity for robust video understanding: equipping MLLMs with external restoration capabilities that can recover degraded evidence before reasoning.

However, effectively using restoration for video understanding is substantially more complex than placing a fixed enhancement model in front of an MLLM. Real-world degradation is often heterogeneous and temporally localized: multiple corruptions may coexist, different failures may occur at different moments, and large portions of a video may remain unaffected. More importantly, the need for recovery is inherently query-dependent. Motion blur may critically obscure evidence for recognizing an action while being irrelevant to a question about scene context; similarly, low illumination may matter only when fine-grained objects or text must be identified. Robust video understanding therefore requires an adaptive mechanism that determines \emph{whether} recovery is necessary, \emph{which} restoration capability should be invoked, and \emph{where} in the video it should be applied.

To address these challenges, we introduce \textbf{ViRA}, a \textbf{Vi}deo \textbf{R}ecovery \textbf{A}gent that casts robust video understanding as an agentic perception process. Rather than treating the degraded video as a fixed input, ViRA actively inspects the visual evidence in the context of the query, decides whether intervention is necessary, and takes recovery actions through external restoration tools. Following a \emph{diagnose--restore--reason} paradigm, the agent determines \emph{whether} recovery is needed, \emph{which} tools to invoke, and \emph{where} in the video each tool should operate; it then observes the recovered visual evidence and performs final reasoning.

Extensive experiments across Video-MME, STAR, and NExT-QA demonstrate that ViRA substantially improves robustness across video durations, corruption patterns, and model backbones. Across different MLLM backbones, ViRA improves over direct inference in diverse benchmark-subset--corruption-regime combinations, yielding an average gain of 3.0 accuracy points. On Qwen3.5-9B, the improvement is consistent across all 15 settings, reaching up to +9.5 points on NExT-QA under full corruption. Notably, these robustness gains do not come at the expense of clean-input performance: on the original uncorrupted videos, ViRA improves the Qwen3.5-9B backbone across all five evaluation subsets, by 2.1 points on average. ViRA is also complementary to query-conditioned frame selection, providing an additional 4.6-point average gain over frame selection alone across three benchmarks and two frame budgets under corruption.


Our key contributions can be summarized as follows:
\begin{itemize}[topsep=3pt,itemsep=3pt,partopsep=0pt]
\item We introduce ViRA, a query-conditioned video recovery agent that actively determines whether recovery is needed, which restoration tools to invoke, and where in the video to apply them before reasoning.
\item We devise a temporally grounded agentic tool-use framework that composes heterogeneous spatial and temporal restoration models through localized recovery actions.
\item Across multiple benchmarks, corruption types, and MLLM backbones, ViRA consistently improves understanding under corrupted inputs while avoiding unnecessary recovery when the evidence is already sufficient.
\end{itemize}

%% file: sec/2_related.tex
\section{Related Work}
\label{sec:related}

\looseness=-1
\textbf{Robust Video Understanding under Corruption}.
Robustness to common corruptions has been systematically benchmarked in image recognition~\cite{hendrycks2019benchmarking}, and subsequently extended to 3D perception and autonomous-driving scenarios~\cite{dong2023benchmarking3d,kong2023robo3d}. As vision systems move toward video understanding, robustness to degraded inputs becomes increasingly important, since real-world footage is frequently affected by blur, noise, compression artifacts, adverse weather, and other spatial-temporal degradations~\cite{safonov2025ntireugc,xie2025star,awad2026awdb}. For instance, Yi \textit{et al.}~\cite{yi2021benchmarking} evaluate action recognition under spatial and temporal corruptions, followed by broader studies across perturbations, occlusions, and temporal action localization~\cite{schiappa2023large,grover2023revealing,zeng2024benchmarking}. More recently, robustness evaluation has expanded to multi-modal models, covering visual--linguistic perturbations~\cite{schiappa2022robustness}, spatio-temporal tampering~\cite{agarwal2025mvtamperbench}, broader trustworthiness~\cite{wang2026benchmarking}, and realistic perturbations for video reasoning in PVRBench~\cite{he2026pvrbench}. On the mitigation side, early approaches rely on corruption-aware training, data augmentation, or test-time adaptation, which improve average accuracy but require anticipating the corruption distribution beforehand. Recent work instead folds robustness into the reasoning process: ROVA~\cite{he2026pvrbench} enforces robustness-aware consistency during training, Robust-R1~\cite{tang2026robustr1} introduces explicit degradation-aware reasoning, and Robust-U1~\cite{tang2026robustu1} equips MLLMs with visual self-recovery prior to reasoning. RestoreAgent~\cite{chen2024restoreagent} and AgenticIR~\cite{zhu2025agenticir} plan restoration sequences for degraded \emph{images}. Robust-TO~\cite{he2026robustto} ranks frames by the product of reliability and query relevance, then routes sub-queries to perception tools such as object detection and action recognition. ViRA instead predicts restoration operators and frame intervals, modifies pixels within those intervals, and returns the recovered video to the MLLM for answering.

\smallskip
\looseness=-1
\noindent\textbf{Video Restoration and Enhancement}.
Visual restoration has developed a rich set of specialized techniques for recovering observations degraded by different physical and digital processes. Classical methods include Richardson--Lucy deconvolution for blur removal~\cite{richardson1972bayesian,lucy1974iterative}, Gray-World color constancy for correcting color shifts~\cite{buchsbaum1980spatial}, the Dark Channel Prior for haze removal~\cite{he2009single}, and CLAHE for local contrast enhancement~\cite{zuiderveld1994contrast}. More recent learning-based approaches provide stronger restoration for challenging degradations: Restormer~\cite{zamir2022restormer} uses efficient transformers for denoising and motion/defocus deblurring, Retinexformer~\cite{cai2023retinexformer} targets low-light enhancement through illumination-guided restoration, and TOENet~\cite{gao2023toenet} jointly addresses haze and sandstorm degradation. Video restoration further exploits temporal redundancy through architectures such as BasicVSR~\cite{chan2021basicvsr} and RVRT~\cite{liang2022rvrt}, while recent all-in-one approaches such as AverNet~\cite{zhao2024avernet} aim to handle multiple unknown and time-varying degradations within a unified model. Despite this progress, restoration is typically applied according to the observed degradation and optimized for perceptual reconstruction quality, independently of the downstream reasoning task. ViRA instead exposes heterogeneous restoration methods as an agentic toolset: conditioned on both the degraded video and the query, it determines \emph{whether} recovery is needed, \emph{which} restoration tools to invoke, and \emph{where} in the video each tool should be applied to recover task-relevant evidence.

%% file: sec/3_method.tex
\section{Method}
\label{sec:method}

We formalize video reasoning under corruption and describe ViRA's recovery pipeline, followed by supervised fine-tuning of the recovery policy.

\subsection{Preliminary}
Let $\mathcal{V}=(I_1,\ldots,I_T)$ be a video with $T$ frames, where $I_t$ is the frame at index $t$, and let $q$ denote a query.
An MLLM $f_{\theta}$, parameterized by $\theta$, takes the video and query as input and generates an answer $\hat{a}=f_{\theta}(\mathcal{V},q)$; we write $\theta_0$ for its  pretrained weights.
The autoregressive generation process can be written as $p_{\theta}(a\mid\mathcal{V},q)=\prod_{i=1}^{|a|}p_{\theta}(a_i\mid a_{<i},\mathcal{V},q)$, where $a_i$ denotes
the $i$-th output token.

In real-world settings, a video may naturally exhibit spatial or temporal degradations, while controlled corruption operators can also be applied to clean videos for systematic robustness evaluation. We denote the degraded observation by $\widetilde{\mathcal{V}}$; for controlled corruptions, $\widetilde{\mathcal{V}}=\mathcal{C}(\mathcal{V})$. A conventional MLLM directly predicts $f_{\theta}(\widetilde{\mathcal{V}},q)$ without explicitly recovering obscured visual evidence. Our goal is to reduce the resulting performance loss by recovering task-relevant evidence before reasoning.

\subsection{ViRA: A Video Recovery Agent}

\textbf{Overview}. We introduce ViRA, which follows a \emph{diagnose--restore--reason} paradigm that integrates visual recovery directly into the video reasoning process. Given a degraded video $\widetilde{\mathcal{V}}$ and a query $q$, ViRA first determines whether the available visual evidence is sufficient for answering the question. If so, it directly produces an answer. Otherwise, it identifies degradations that obscure task-relevant evidence, localizes the affected temporal regions, and generates one or more restoration tool calls. Importantly, this decision is conditioned on the downstream query: a visible degradation does not necessarily require recovery if the information needed to answer the query remains accessible.

\smallskip
\noindent
\textbf{Agentic Tool Invocation}. Let $\mathcal{T}=\{\tau_1,\tau_2,\ldots,\tau_K\}$ denote a library of restoration tools, covering spatial and temporal degradations such as low illumination, blur, noise, haze, and color distortion. Rather than assigning one operation to the entire video, ViRA selects tools and their temporal regions jointly. The $j$-th temporally grounded tool call is
\begin{equation}
u_j=\left(\tau_{k_j},[l_j,r_j]\right),
\quad
\tau_{k_j}\in\mathcal{T},
\quad
1\leq l_j\leq r_j\leq T,
\end{equation}
where $[l_j,r_j]$ is the tool's target frame range.

To formalize the agent action space, let
\begin{equation}
\mathcal{S}
=
\left\{
[l,r]\mid l,r\in\mathbb{Z},\;1\leq l\leq r\leq T
\right\}
\end{equation}
denote the set of valid temporal intervals. A recovery action consists of an ordered sequence of tool--interval pairs,
\begin{equation}
\mathcal{U}_m
=
\left(
u_1,u_2,\ldots,u_m
\right)
\in
\left(\mathcal{T}\times\mathcal{S}\right)^m,
\end{equation}
where $m\geq1$ denotes the number of tool calls. Conditioned on the video, query, and tool descriptions, ViRA decodes
\begin{equation}
z=\operatorname{Decode}\!\left[p_{\theta}(\cdot\mid\widetilde{\mathcal{V}},q,\mathcal{T})\right].
\end{equation}
Here, decoding is deterministic, and $z$ is either $(\textsc{Answer},a)$, containing a direct answer, or a recovery plan $\mathcal{U}_m$. Calls may reuse tools and target different or overlapping intervals. This temporally grounded action space lets ViRA selectively recover query-relevant evidence rather than uniformly restoring the entire video.

\smallskip
\noindent
\textbf{Tool Execution and Video Recovery}. Once ViRA predicts a recovery action $\mathcal{U}_m$, the tool calls are executed sequentially. Starting from $\mathcal{V}^{(0)}=\widetilde{\mathcal{V}}$, the $j$-th call applies $\tau_{k_j}$ only to its designated temporal interval:
\begin{equation}
\mathcal{V}^{(j)}_{l_j:r_j}
=
\tau_{k_j}
\left(
\mathcal{V}^{(j-1)}_{l_j:r_j}
\right),
\qquad
j=1,\ldots,m,
\end{equation}
while frames outside the interval remain unchanged,
\begin{equation}
\mathcal{V}^{(j)}_{t}
=
\mathcal{V}^{(j-1)}_{t},
\qquad
t\notin[l_j,r_j].
\end{equation}
After all predicted calls have been executed, the recovered video is defined as
\begin{equation}
\mathcal{V}^{\mathrm{rec}}
=
\mathcal{V}^{(m)}
=
\mathcal{R}
\left(
\widetilde{\mathcal{V}}; \,\mathcal{U}_m
\right),
\end{equation}
where $\mathcal{R}$ denotes sequential execution, including composition on overlapping intervals. Localized execution supports temporally varying corruptions and compound degradations: different tools can address different regions, while successive calls can refine the same region without changing frames outside their selected intervals.

\smallskip
\noindent
\textbf{Final Reasoning}. The no-recovery branch returns the answer from the first output; only the recovery branch requires another MLLM call:
\begin{equation}
\hat{a}=
\begin{cases}
a, & z=(\textsc{Answer},a),\\
f_{\theta}(\mathcal{V}^{\mathrm{rec}},q), & z=\mathcal{U}_m.
\end{cases}
\end{equation}
ViRA uses a single decision stage: it jointly determines whether intervention is necessary, which tools to invoke, and where to apply them. Once the plan is executed, the model reasons over the recovered evidence without another recovery decision. Algorithm~\ref{alg:vira} summarizes the procedure.

\smallskip
\noindent
\textbf{Zero-shot and Fine-tuned ViRA Variants}. 
Our diagnose--restore--reason framework requires no architectural modification: the MLLM directly serves as the recovery policy $p_{\theta}(z\mid\widetilde{\mathcal{V}},q,\mathcal{T})$, conditioned on the query, video, and natural-language tool descriptions. We evaluate two variants with the same backbone, tool library, and inference procedure. ViRA-ZS uses the pretrained weights $\theta_0$ without training, while ViRA uses the supervised fine-tuned (SFT) weights $\theta_{\mathrm{SFT}}$ learned in \S\ref{subsec:sft}. In both cases, the same MLLM is used for diagnosis and final reasoning, while all restoration tools remain frozen. ViRA-ZS measures the benefit of agentic recovery alone, whereas ViRA further captures the gain from learning the recovery policy.

\begin{algorithm}[t]
\caption{ViRA inference: diagnose--restore--reason}
\label{alg:vira}
\begin{algorithmic}[1]
\Require degraded video $\widetilde{\mathcal{V}}$, query $q$, tool library $\mathcal{T}$, video MLLM $f_{\theta}$ with $\theta\in\{\theta_0,\theta_{\mathrm{SFT}}\}$
\Ensure answer $\hat{a}$
\State $z\gets\operatorname{Decode}[p_{\theta}(\cdot\mid\widetilde{\mathcal{V}},q,\mathcal{T})]$
\If{$z=(\textsc{Answer},a)$}
    \State \Return $a$ \Comment{No second MLLM call}
\Else \Comment{Restore: $z=\mathcal{U}_m=(u_1,\ldots,u_m)$}
    \State $\mathcal{V}^{(0)}\gets\widetilde{\mathcal{V}}$
    \For{$j=1,\ldots,m$}
        \State $(\tau_{k_j},[l_j,r_j])\gets u_j$
        \State $\mathcal{V}^{(j)}\gets\mathcal{V}^{(j-1)}$ \Comment{outside $[l_j,r_j]$ unchanged}
        \State $\mathcal{V}^{(j)}_{l_j:r_j}\gets\tau_{k_j}\big(\mathcal{V}^{(j-1)}_{l_j:r_j}\big)$
    \EndFor
    \State $\mathcal{V}^{\mathrm{rec}}\gets\mathcal{V}^{(m)}$
\EndIf
\State $\hat{a}\gets f_{\theta}(\mathcal{V}^{\mathrm{rec}},q)$ \Comment{Reason}
\State \Return $\hat{a}$
\end{algorithmic}
\end{algorithm}

\subsection{Supervised Fine-tuning for Better Recovery}
\label{subsec:sft}
Although ViRA-ZS already benefits from the agentic recovery procedure, its zero-shot recovery decisions are often unreliable: the model may fail to identify task-relevant degradations, invoke unnecessary restoration tools, select an inappropriate tool, or localize the corrupted temporal region incorrectly. We therefore apply supervised fine-tuning (SFT) to $\theta$ to improve the recovery policy, yielding ViRA.


\smallskip
\noindent
\textbf{Training Data}. For each training example, we construct a degraded video $\widetilde{\mathcal{V}}_i$, a query $q_i$, and a target agent response $y_i$. The SFT dataset is written as
\begin{equation}
\mathcal{D}_{\mathrm{SFT}}
=
\{(\widetilde{\mathcal{V}}_i,q_i,y_i)\}_{i=1}^{N}.
\end{equation}
The target $y_i$ specifies the desired agent behavior: it contains a direct answer when the degraded video provides sufficient evidence, or a recovery plan $\mathcal{U}_m$ followed by the final answer when restoration is needed. Each tool call identifies the restoration tool and its target frame interval. A recovery plan can contain multiple calls when different degradations or temporal regions require intervention. The visual input to SFT is always the degraded video; recovered frames are used to verify the recovery target during data construction, not as training inputs.

\smallskip
\noindent
\textbf{Target Construction}. The supervision targets $y_i$ are generated
automatically by the backbone itself, without human annotation or distillation
from a stronger model. For each (question, corruption, layout) cell we sample
the frame budget once, apply the corruption, and query the frozen backbone
$f_{\theta_0}$ twice under deterministic decoding: once with the original
option order, and once with a seeded non-identity permutation of the options.
If both queries are correct, the degraded evidence is already sufficient and we
emit an \textsc{Answer} target. If both are wrong, we apply the restoration
tool mapped to that corruption and query again in both option orders; if both
are now correct, we emit a \textsc{Recover} target containing the tool call
and verified answer. All remaining cases are
discarded. The two-order agreement check reduces chance-correct supervision,
and recovery is never attempted on cells the
model already answers correctly, so the policy is not trained to restore
unnecessarily. Each corruption type is mapped to exactly one tool, and
\emph{low-light noise} is excluded from extraction entirely, so it remains an
unseen degradation at evaluation time. Recovery plans specify only tool
identities and frame indices; no chain-of-thought is generated or stored.

\smallskip
\noindent
\textbf{Supervised Fine-tuning}. Given the degraded video, query, and available restoration tools, ViRA is trained to predict the target response $y_i$. We optimize the standard autoregressive language modeling objective
\begin{equation}
\mathcal{L}_{\mathrm{SFT}}
=
-
\sum_{i=1}^{N}
\sum_{t=1}^{|y_i|}
\log
p_{\theta}
(
y_{i,t}
\mid
y_{i,<t},
\widetilde{\mathcal{V}}_i,
q_i,
\mathcal{T}
).
\end{equation}
Here, $y_{i,t}$ is the $t$-th target token, including both tool-call and answer tokens for recovery examples. Starting from $\theta_0$, we train LoRA adapters in the language layers while freezing the backbone and restoration tools. The resulting effective MLLM parameters are denoted by $\theta_{\mathrm{SFT}}$. Explicit tool identities and frame ranges supervise which tool to use and where to apply it, while direct-answer examples teach the policy to avoid recovery when evidence is sufficient.


%% file: sec/4_experiments.tex
\section{Experiments}
\label{sec:exp}

We evaluate ViRA on corrupted video understanding benchmarks and naturally degraded driving videos.

\subsection{Experimental Settings}
\textbf{Models and Benchmarks}.
We evaluate ViRA with four MLLM backbones: Qwen3.5-9B~\cite{qwen35}, Qwen3-VL-8B~\cite{bai2025qwen3}, LLaVA-OV2-8B~\cite{an2026llava} and GPT-5.4. For Qwen models, ViRA augments the frozen backbone with a rank-16 LoRA adapter. Since GPT-5.4 does not support parameter-efficient fine-tuning in our setting, we evaluate only its zero-shot agentic variant, ViRA-ZS.
We conduct extensive experiments on four video question answering benchmarks and a set of real-world driving videos:

\begin{itemize}
\item \textbf{Video-MME~\cite{fu2025videomme}} is a comprehensive benchmark for video understanding across different video durations. We evaluate its Short, Medium, and Long subsets, enabling us to study robustness as temporal context increases.
\item \textbf{STAR~\cite{wu2021star}} is a multiple-choice benchmark for situated reasoning in videos, covering interaction, sequence, prediction, and feasibility questions that require understanding temporal dynamics and human--object interactions.
\item \textbf{NExT-QA~\cite{xiao2021next}} is a five-choice benchmark emphasizing temporal and causal reasoning beyond descriptive perception. We use the official validation split, which is disjoint from the training data used to construct our SFT set.
\item \textbf{VRU-Accident~\cite{kim2025vruaccident}} is a vision-language benchmark for accident scene understanding involving vulnerable road users. We use it only for qualitative case studies to examine ViRA's recovery and reasoning behavior in safety-critical driving scenarios.
\item \textbf{Real-world driving videos} are an evaluation set of 253 driving videos with 759 multiple-choice questions that we curate to assess whether recovery transfers beyond controlled corruptions. We select these videos from a collection assembled from DrivingDojo~\cite{wang2024drivingdojo}, Boreas~\cite{burnett2023boreas}, Boreas-RT~\cite{lisus2026boreasrt}, BDD100K~\cite{yu2020bdd100k}, and DrivingHazy/GoProHazy~\cite{fan2024dvd}.  Note that the visual degradations occur naturally in the recorded footage; no synthetic corruption is added. 
\end{itemize}

\input{tables/main_table}

\smallskip
\noindent\textbf{Evaluation Protocol}.
We construct controlled corruptions at a fixed severity using 10 degradation types and evaluate three corruption regimes: \emph{All}, where a single corruption type is applied to all 16 frames; \emph{Half}, where a seeded contiguous block of 8 out of 16 frames is corrupted; and \emph{Mixed}, where four distinct corruption types are applied to four contiguous 4-frame blocks. For each benchmark and backbone, we evaluate four settings: \emph{Clean}, which uses the original uncorrupted video; \emph{No recovery}, which directly answers from the corrupted input without recovery; \emph{ViRA-ZS}, which uses the frozen backbone as a zero-shot recovery agent under our tool-use protocol; and \emph{ViRA}, which uses the learned recovery policy. We evaluate the learned policy ViRA on the two open-weight Qwen backbones. GPT-5.4 is evaluated only under \emph{Clean}, \emph{No recovery}, and \emph{ViRA-ZS}, as its parameters are not available for training; LLaVA-OV2-8B is likewise reported under these three settings only. For comparison with degradation-aware reasoning, we re-implement Robust-R1~\cite{tang2026robustr1} for video QA and evaluate its SFT and RL variants alongside the Qwen2.5-VL-3B base model on our real-world set (Table~\ref{tab:realworld}).

\smallskip
\noindent
\textbf{Degradation Types}.
We evaluate 10 degradation types in four categories.
(i) \emph{Exposure and illumination}: \emph{underexposure}, \emph{overexposure}, and \emph{low-light noise}, where poor illumination is accompanied by amplified sensor noise.
(ii) \emph{Color distortion}: \emph{color cast}, which shifts the global color balance, and \emph{chromatic aberration}, which produces wavelength-dependent color fringing near object boundaries.
(iii) \emph{Optical blur}: \emph{defocus blur} and \emph{motion blur}, caused by incorrect focus and camera or object motion, respectively.
(iv) \emph{Atmospheric effects and occlusion}: \emph{fog/smoke}, \emph{dust storm}, and \emph{glass occlusion}.
The first seven degradations use parameterized corruption operators $\mathcal{C}$, whereas the latter three are constructed from real captured degradation patterns, introducing spatially non-uniform and temporally varying effects difficult to capture with simple parametric operators.

\smallskip
\noindent\textbf{Restoration Models}. 
We use a diverse set of restoration methods in our recovery toolbox: RetinexFormer~\cite{cai2023retinexformer} enhances frames degraded by underexposure. TOENet~\cite{gao2023toenet} restores visibility under dust- and sandstorm-like degradations and glass occlusion. Restormer~\cite{zamir2022restormer} is used for motion deblurring, while Richardson--Lucy deconvolution~\cite{richardson1972bayesian,lucy1974iterative} is applied to defocus blur. Gray-World color constancy~\cite{buchsbaum1980spatial} corrects global color shifts, and channel realignment is used to mitigate chromatic aberration. For haze and fog, we apply the Dark Channel Prior~\cite{he2009single}; for other low-contrast degradations, we use CLAHE~\cite{zuiderveld1994contrast} to improve local contrast. Note that all restoration modules are classical or regression-based rather than generative, reducing the risk of introducing visually plausible but unsupported evidence into downstream reasoning.

\smallskip
\noindent\textbf{Implementation Details}.
We train ViRA by adapting only the language layers of the backbone with a rank-16 LoRA adapter ($r{=}16$, $\alpha{=}32$, dropout $0.05$), while keeping the remaining model parameters frozen. The SFT corpus contains 13{,}729 examples, including 10{,}635 full-corruption recovery examples, 1{,}075 half-corruption examples, 646 mixed-corruption examples requiring multiple recovery actions, and 1{,}373 direct-answer examples for cases where recovery is unnecessary. 
At inference, the model either answers directly or emits recovery actions under the executable tool schema in Section~\ref{sec:method}. Selected tools operate sequentially on the requested frames, and the recovered video is returned to the MLLM for final prediction. We use deterministic decoding for tool selection and constrained decoding for multiple-choice answers. All models receive the same 16 sampled frames and identical corrupted inputs for each evaluation instance unless otherwise specified.

\begin{figure*}[t]
\centering

\includegraphics[width=\textwidth]{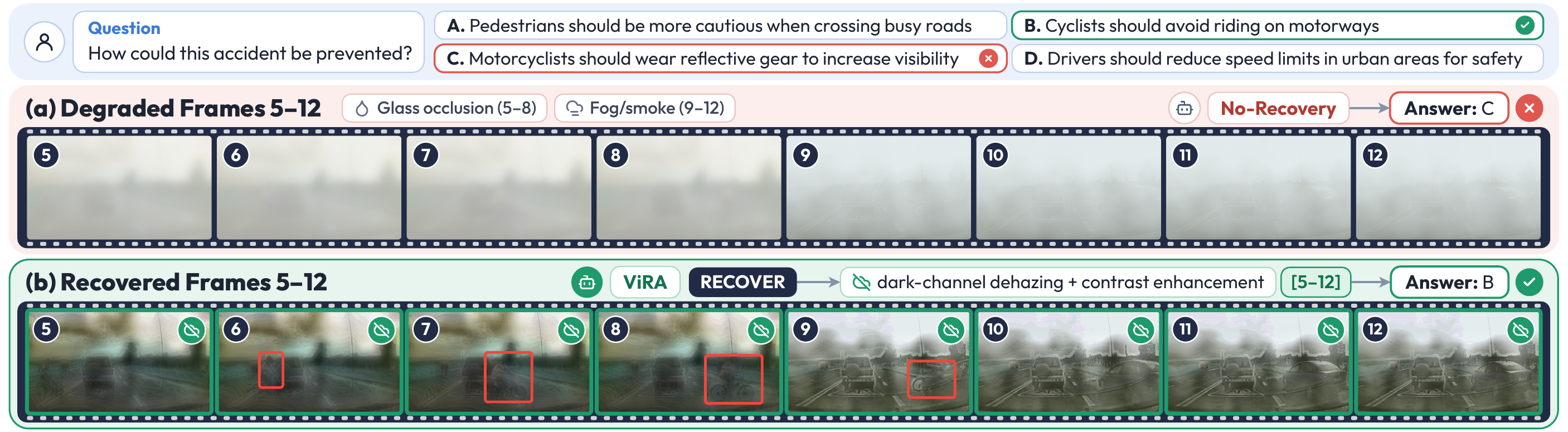}
\vspace{-7pt}
\caption{
\textbf{Qualitative case study on VRU-Accident under mixed corruptions.}
The 16-frame input contains four corruption blocks: overexposure (1--4),
glass occlusion (5--8), fog/smoke (9--12), and underexposure (13--16).
Direct inference predicts the incorrect answer
\textcolor{wrongred}{\textbf{C}}, while ViRA localizes the task-relevant
evidence to frames 5--12 and selectively applies recovery.
The restored view makes the cyclist entering the motorway and the subsequent
collision more discernible, leading to the correct answer
\textcolor{correctgreen}{\textbf{B}}.}
\label{fig:vru_case}
\vspace{-10pt}
\end{figure*}

\subsection{Experimental Results}
\noindent
\textbf{Main Results}.
Table~\ref{tab:main} reports results across three benchmarks, three corruption regimes, and four MLLM backbones.
Visual corruption substantially degrades direct video understanding, particularly when task-relevant evidence is affected throughout the video.
On Qwen3.5-9B, ViRA improves over \emph{No recovery} by 5.0 points on average under the all-frame regime (A), with gains of up to 9.5 points on NExT-QA.
The improvement remains substantial under the mixed regime (M), averaging 4.3 points despite the presence of heterogeneous degradations within the same video.
Even when only half of the video is corrupted (H), ViRA provides a further 1.7-point average gain, showing that the learned policy can selectively recover useful evidence without requiring the entire video to be degraded.

On Qwen3-VL-8B, ViRA improves the all-frame and mixed regimes by 3.9 and 2.6 points on average, respectively, including gains of 7.9 points on NExT-QA under full corruption and 5.1 points under mixed corruption.
Across both Qwen backbones, the strongest improvements consistently appear under the more challenging A and M settings, where corrupted evidence is difficult to bypass through unaffected frames.
ViRA-ZS also provides meaningful recovery without task-specific adaptation, while the learned policy generally yields larger and more reliable gains.
Moreover, on the API-only GPT-5.4, ViRA-ZS improves all five full-corruption subsets, with gains of up to 4.3 points, showing that agentic recovery can benefit proprietary MLLMs without access to their parameters.

\input{tables/realworld_table}

\smallskip
\noindent
\textbf{Transfer to Real-World Degradations.}
On the real-world driving videos (Table~\ref{tab:realworld}), ViRA improves accuracy from 78.7\% to 81.2\% on Qwen3.5-9B and from 74.4\% to 75.6\% on Qwen3-VL-8B, gains of 2.5 and 1.2 percentage points, which supports the transfer of its learned recovery policy from controlled corruptions to naturally degraded videos.
ViRA-ZS yields mixed results, improving LLaVA-OV2-8B by 0.8 points but reducing accuracy on both Qwen backbones.
On Qwen2.5-VL-3B, neither re-implemented Robust-R1 variant improves over the base model on this set: they score 63.1\% (SFT) and 62.7\% (RL), against 65.1\%.
\par

\begin{figure*}[t]
    \centering
    \begin{subfigure}[t]{0.32\textwidth}
        \centering
        \includegraphics[width=\linewidth]{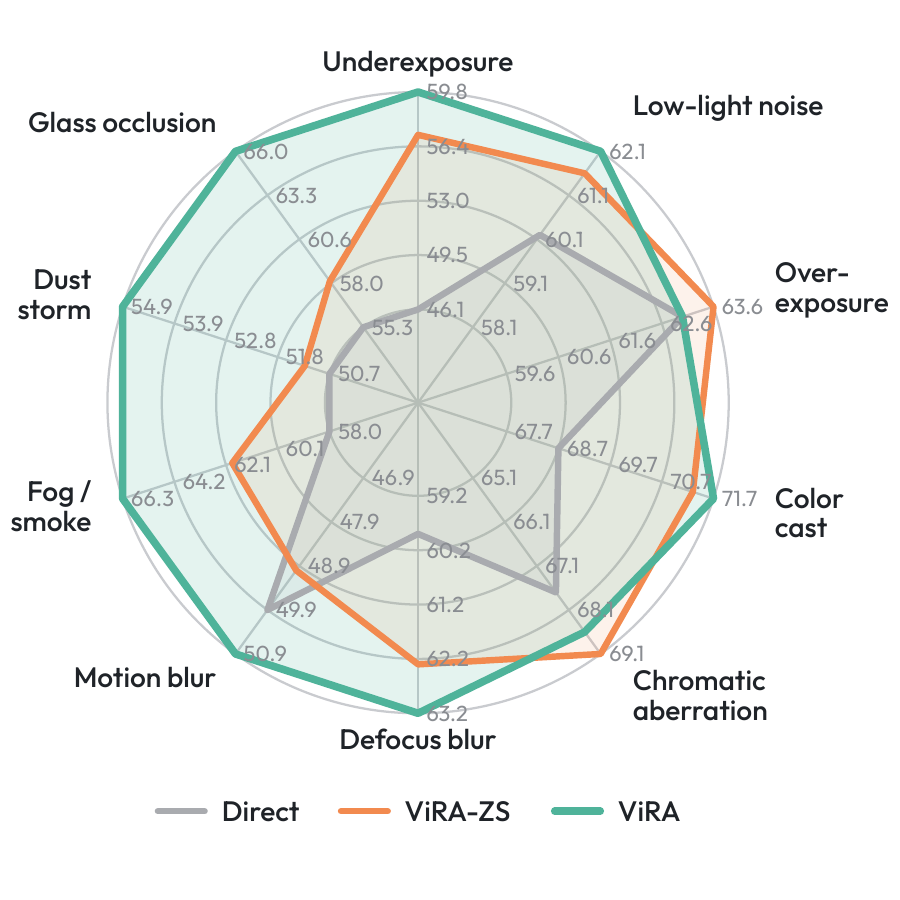}
        \vspace{-20pt}
        \caption{Video-MME Short}
        \label{fig:radar_short}
    \end{subfigure}
    \hfill
    \begin{subfigure}[t]{0.32\textwidth}
        \centering
        \includegraphics[width=\linewidth]{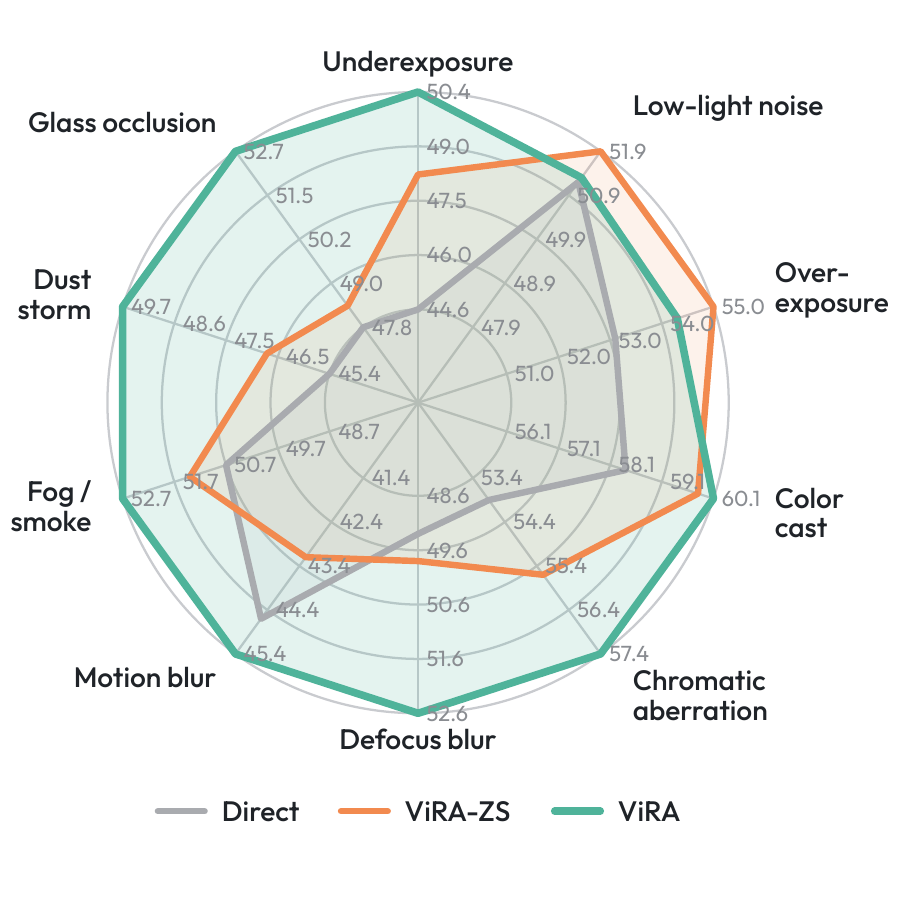}
        \vspace{-20pt}
        \caption{Video-MME Medium}
        \label{fig:radar_medium}
    \end{subfigure}
    \hfill
    \begin{subfigure}[t]{0.32\textwidth}
        \centering
        \includegraphics[width=\linewidth]{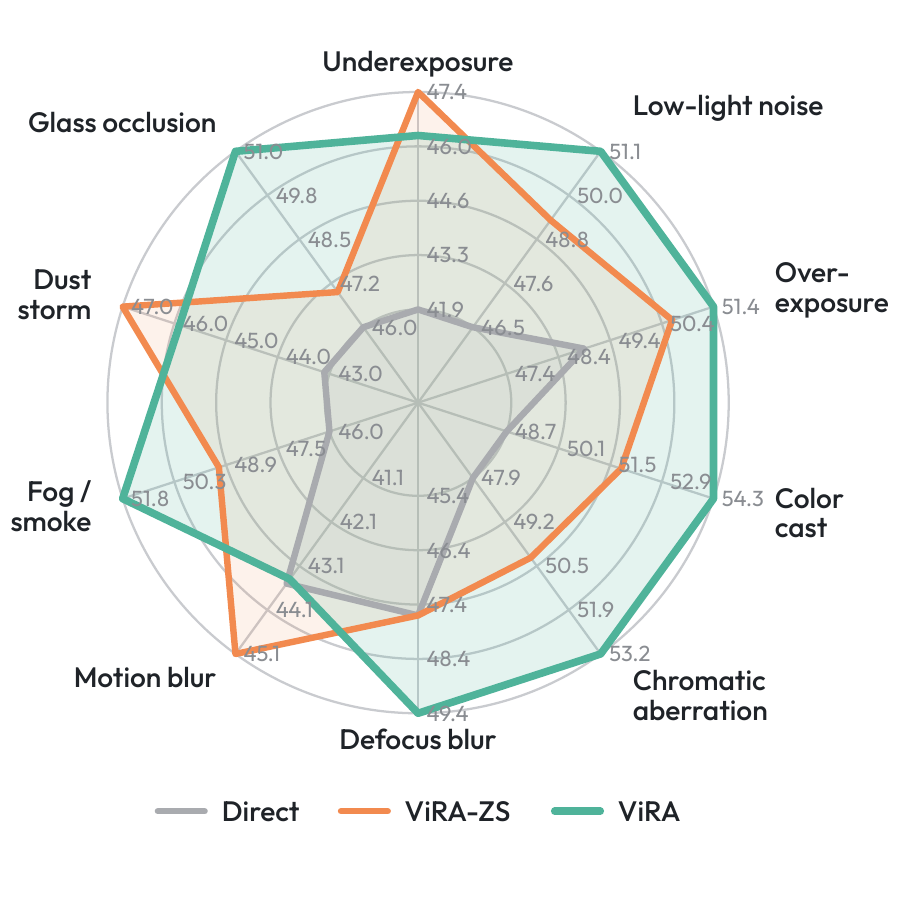}
        \vspace{-20pt}
        \caption{Video-MME Long}
        \label{fig:radar_long}
    \end{subfigure}
    \vspace{-5pt}
    \caption{
    \textbf{Per-degradation robustness on Video-MME.}
    Accuracy (\%) of Qwen3.5-9B with Direct inference, ViRA-ZS, and ViRA under the full-corruption (\emph{All}) regime on
    (a) short-, (b) medium-, and (c) long-form videos.
    Each axis uses its own accuracy range, shown by the tick labels; compare methods within an axis.
    }
    \label{fig:per_degradation}
    \vspace{-10pt}
\end{figure*}

\begin{figure}[t]
    \centering
    \includegraphics[width=\linewidth]{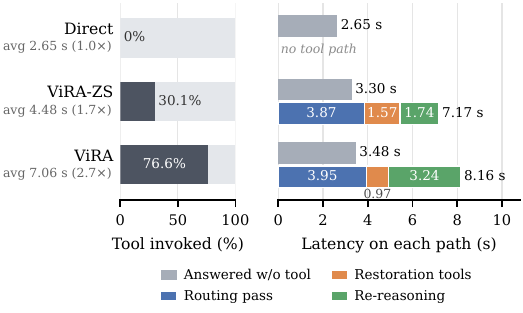}
    \vspace{-20pt}
    \caption{
    \textbf{Efficiency analysis on fully corrupted videos.}
    Left: fraction of samples that invoke at least one restoration tool.
    Right: average latency of the no-recovery path (gray) and recovery path, decomposed into routing, restoration, and re-reasoning stages.
    Numbers beside each method report the overall average latency and its relative cost compared with direct Qwen3.5-9B inference.
    }
    \vspace{-5pt}
    \label{fig:efficiency}
\end{figure}

\input{tables/clean_table}

\smallskip
\noindent\textbf{Results on Clean Videos}.
A practical recovery agent should preserve performance when visual restoration is unnecessary. We therefore also evaluate the original clean videos. Table~\ref{tab:clean} shows that both ViRA-ZS and ViRA outperform the base Qwen3.5-9B on all five subsets, with ViRA performing best overall and gaining 2.1 points on average. Nominally ``clean'' videos may still contain low illumination, blur, or compression artifacts. We speculate that selectively recovering such imperfect evidence can improve the visual signal available for reasoning, contributing to the gains even without injected corruption.

\smallskip
\noindent\textbf{Qualitative Study on Autonomous Driving.}
In Figure~\ref{fig:vru_case}, mixed corruptions obscure how the accident unfolds, causing direct inference to fail. ViRA selectively restores the relevant temporal window, making the cyclist's motorway entry and subsequent collision more discernible and yielding the correct prevention-oriented answer. The selected window covers frames 5--12, spanning glass occlusion and fog/smoke, while leaving the other two corruption blocks unchanged. This case illustrates how recovery can target the evidence needed to answer the question without restoring every degraded frame.

\subsection{Further Discussions}

\noindent\textbf{Per-Degradation Results}.
Figure~\ref{fig:per_degradation} breaks down robustness across individual corruption types on Video-MME.
ViRA consistently expands the robustness profile of the direct backbone across short-, medium-, and long-form videos, with gains spanning illumination, color, blur, and visibility degradations.
Notably, ViRA-ZS already improves robustness to many corruption types without any task-specific training, while the fine-tuned ViRA generally provides broader and more stable improvements across degradations.
Gains across all three durations suggest that the policy is not specialized to a single temporal regime.


\smallskip
\noindent\textbf{Efficiency Analysis}.
Figure~\ref{fig:efficiency} breaks down agentic recovery costs.
Direct inference takes 2.65\,s per sample on average, while ViRA-ZS and ViRA take 4.48\,s ($1.7\times$) and 7.06\,s ($2.7\times$).
This overhead is adaptive rather than uniform: a restoration tool is invoked on only 30.1\% of samples for ViRA-ZS and 76.6\% for ViRA (left), the latter reflecting the more aggressive recovery policy learned under corrupted inputs.
When no tool is invoked, the agent adds only 0.65--0.83\,s over direct inference (3.30\,s and 3.48\,s; right, gray), the fixed cost of the routing decision.
When a tool is invoked, latency rises to 7.17\,s and 8.16\,s, but restoration itself accounts for a small fraction of it; the rest is MLLM inference for routing ($\approx$3.9\,s) and re-reasoning over the recovered frames (1.74\,s and 3.24\,s).
These measurements use fully corrupted inputs. If cleaner inputs trigger recovery less often, average latency can approach the no-tool path (3.3--3.5\,s, about $1.3\times$ direct inference), concentrating extra computation on samples requiring recovery.

\smallskip
\noindent\textbf{Adaptive Keyframe Sampling}.
Adaptive keyframe sampling reduces redundant visual input by selecting frames relevant to the query. We combine ViRA with AKS~\cite{tang2025adaptive} to assess whether visual recovery provides additional benefits under limited frame budgets.
Table~\ref{tab:aks} shows that AKS alone offers limited gains over uniform sampling on corrupted videos. Adding ViRA improves accuracy over AKS alone by 3.8--6.0 percentage points across all three benchmarks and both frame budgets, averaging 4.6 points across the six settings.
These gains show that adaptive sampling and visual recovery are complementary: selecting relevant moments does not eliminate the need to recover degraded evidence within them.

\begin{table}[t]
\centering
\caption{
\textbf{Compatibility with adaptive key frame sampling (AKS).}
Accuracy (\%) of Qwen3.5-9B on Video-MME Short, STAR, and NExT-QA
under 8- and 16-frame budgets. 
We evaluate AKS with ViRA-ZS and ViRA under corrupted inputs. Corrupted inputs use the full-corruption layout.
}
\label{tab:aks}
\small
\setlength{\tabcolsep}{6pt}

\resizebox{\columnwidth}{!}{%
\begin{tabular}{lcccccc}
\toprule
\multirow{2}{*}{\textbf{Method}}
& \multicolumn{2}{c}{\textbf{V-MME Short}}
& \multicolumn{2}{c}{\textbf{STAR}}
& \multicolumn{2}{c}{\textbf{NExT-QA}} \\
\cmidrule(lr){2-3}
\cmidrule(lr){4-5}
\cmidrule(lr){6-7}
& \textbf{8f} & \textbf{16f}
& \textbf{8f} & \textbf{16f}
& \textbf{8f} & \textbf{16f} \\
\midrule

\g{Clean (ref.)}
& \g{67.9} & \g{71.8}
& \g{68.3} & \g{69.0}
& \g{79.0} & \g{81.8} \\


Uniform
& 54.4 & 57.9
& 46.3 & 47.7
& 65.2 & 65.6 \\
AKS
& 56.9 & 59.3
& 46.7 & 47.7
& 64.9 & 66.9 \\
AKS + ViRA-ZS
& 57.1 & 60.4
& 44.6 & 47.6
& 68.0 & 68.1 \\
AKS + ViRA
& \textbf{61.2} & \textbf{63.1}
& \textbf{52.7} & \textbf{52.2}
& \textbf{69.7} & \textbf{71.3} \\

\bottomrule
\end{tabular}%
}
 \vspace{-5pt}
\end{table}



%% file: tables/main_table.tex
\begin{table*}[t]
\centering

\caption{
\textbf{Robust video understanding under visual corruption.}
Accuracy (\%) on three benchmarks with four MLLM backbones.
Corrupted inputs are evaluated under three regimes:
\textit{A} (All) corrupts every frame,
\textit{H} (Half) corrupts half of the video,
and \textit{M} (Mixed) applies four distinct corruption types to four consecutive 4-frame blocks.
\textit{Clean (ref.)} evaluates the uncorrupted videos and provides a clean-input reference.
\textit{No recovery} feeds corrupted videos directly to the MLLM;
\textit{ViRA-ZS} is our zero-shot recovery agent and \textit{ViRA} our learned recovery policy.
Within each backbone, the best result among corrupted settings is in \textbf{bold} and the second best is \underline{underlined}.
}
\vspace{-5pt}
\label{tab:main}
\footnotesize
\setlength{\tabcolsep}{6pt}
\resizebox{\linewidth}{!}{
\begin{tabular}{ll ccc ccc ccc ccc ccc}
\toprule
\multirow{3}{*}{\textbf{Model}} & \multirow{3}{*}{\textbf{Method}}
  & \multicolumn{9}{c}{\textbf{Video-MME~\cite{fu2025videomme}}}
  & \multicolumn{3}{c}{\multirow{2}{*}{\textbf{STAR~\cite{wu2021star}}}}
  & \multicolumn{3}{c}{\multirow{2}{*}{\textbf{NExT-QA~\cite{xiao2021next}}}} \\
\cmidrule(lr){3-11}
& & \multicolumn{3}{c}{\textbf{Short}} & \multicolumn{3}{c}{\textbf{Medium}} & \multicolumn{3}{c}{\textbf{Long}}
  & \multicolumn{3}{c}{} & \multicolumn{3}{c}{} \\
\cmidrule(lr){3-5}\cmidrule(lr){6-8}\cmidrule(lr){9-11}\cmidrule(lr){12-14}\cmidrule(lr){15-17}
& & A & H & M & A & H & M & A & H & M & A & H & M & A & H & M \\
\midrule
\multirow{4}{*}{Qwen3.5-9B~\cite{qwen35}}
& \g{Clean (ref.)}
& \multicolumn{3}{c}{\g{71.8}} & \multicolumn{3}{c}{\g{58.6}} & \multicolumn{3}{c}{\g{53.0}} & \multicolumn{3}{c}{\g{69.0}} & \multicolumn{3}{c}{\g{81.8}} \\
& No recovery
& 57.9 & 67.7 & 60.6 & 49.9 & 56.7 & 50.8 & 46.7 & 50.5 & 46.4 & 47.7 & 65.9 & 53.4 & 65.6 & 78.8 & 71.4 \\
& ViRA-ZS
& \underline{60.5} & \underline{69.3} & \underline{62.2} & \underline{51.0} & \textbf{58.5} & \underline{51.1} & \underline{49.2} & \textbf{53.2} & \underline{49.7} & \underline{49.7} & \underline{66.4} & \underline{53.9} & \underline{67.0} & \underline{80.1} & \underline{73.5} \\
& ViRA
& \textbf{62.6} & \textbf{69.7} & \textbf{64.4} & \textbf{52.7} & \underline{57.3} & \textbf{52.8} & \textbf{49.8} & \underline{52.1} & \textbf{50.5} & \textbf{52.6} & \textbf{67.2} & \textbf{57.6} & \textbf{75.1} & \textbf{81.9} & \textbf{78.7} \\
\midrule
\multirow{4}{*}{Qwen3-VL-8B~\cite{bai2025qwen3}}
& \g{Clean (ref.)}
& \multicolumn{3}{c}{\g{71.7}} & \multicolumn{3}{c}{\g{57.7}} & \multicolumn{3}{c}{\g{50.6}} & \multicolumn{3}{c}{\g{70.6}} & \multicolumn{3}{c}{\g{80.8}} \\
& No recovery
& 56.9 & \underline{67.5} & 60.1 & 47.5 & \textbf{54.7} & 50.0 & 46.7 & 49.3 & \underline{48.0} & 48.9 & \underline{66.9} & \underline{55.7} & 66.9 & \underline{79.2} & 73.3 \\
& ViRA-ZS
& \underline{60.5} & 65.1 & \underline{62.4} & \underline{49.1} & 52.7 & \underline{50.3} & \underline{47.5} & \textbf{51.5} & 45.7 & \textbf{52.3} & \textbf{67.4} & 54.1 & \underline{67.5} & 77.7 & \underline{74.6} \\
& ViRA
& \textbf{62.1} & \textbf{69.2} & \textbf{64.2} & \textbf{49.5} & \underline{53.6} & \textbf{50.6} & \textbf{48.3} & \underline{49.8} & \textbf{50.1} & \underline{51.9} & 65.8 & \textbf{56.8} & \textbf{74.8} & \textbf{81.3} & \textbf{78.4} \\
\midrule
\multirow{3}{*}{LLaVA-OV2-8B~\cite{an2026llava}}
& \g{Clean (ref.)}
& \multicolumn{3}{c}{\g{71.7}} & \multicolumn{3}{c}{\g{57.8}} & \multicolumn{3}{c}{\g{50.9}} & \multicolumn{3}{c}{\g{67.9}} & \multicolumn{3}{c}{\g{81.1}} \\
& No recovery
& \underline{58.4} & \textbf{67.6} & \underline{60.3} & \underline{48.3} & \textbf{55.1} & \underline{50.4} & \textbf{45.8} & \underline{49.2} & \textbf{47.9} & \underline{49.1} & \textbf{65.2} & \underline{54.8} & \underline{67.3} & \textbf{78.8} & \underline{74.0} \\
& ViRA-ZS
& \textbf{59.5} & \underline{67.4} & \textbf{60.8} & \textbf{49.7} & \underline{54.3} & \textbf{50.6} & \underline{45.0} & \textbf{49.7} & \underline{46.6} & \textbf{50.4} & \underline{64.9} & \textbf{55.4} & \textbf{70.0} & \underline{78.6} & \textbf{74.4} \\
\midrule
\multirow{3}{*}{GPT-5.4}
& \g{Clean (ref.)}
& \multicolumn{3}{c}{\g{79.4}} & \multicolumn{3}{c}{\g{66.4}} & \multicolumn{3}{c}{\g{62.4}} & \multicolumn{3}{c}{\g{69.9}} & \multicolumn{3}{c}{\g{83.3}} \\
& No recovery
& \underline{63.0} & \underline{74.0} & \underline{67.0} & \underline{58.8} & \underline{64.8} & \textbf{59.6} & \underline{57.0} & \underline{60.7} & \underline{57.0} & \underline{47.6} & \textbf{66.7} & \underline{53.2} & \underline{70.4} & \underline{81.0} & \textbf{75.9} \\
& ViRA-ZS
& \textbf{66.5} & \textbf{74.4} & \textbf{67.2} & \textbf{60.4} & \textbf{65.1} & \underline{58.0} & \textbf{57.3} & \textbf{60.7} & \textbf{57.2} & \textbf{51.8} & \textbf{66.7} & \textbf{55.6} & \textbf{74.7} & \textbf{81.0} & \underline{75.7} \\
\bottomrule
\end{tabular}
}
\vspace{-5pt}
\end{table*}

%% file: tables/realworld_table.tex
\begin{table}[t]
\centering
\caption{\textbf{Transfer to naturally degraded driving videos.}
Accuracy (\%) on 253 real-world driving videos with 759 questions.
Top: ViRA variants. Bottom: our video-QA re-implementation of Robust-R1~\cite{tang2026robustr1}.
Subscripts give the change relative to the same backbone without recovery; each backbone's best result is in \textbf{bold}.}
\label{tab:realworld}
\footnotesize
\setlength{\tabcolsep}{0pt}
\renewcommand{\arraystretch}{1.12}
\newcommand{\rwd}[3]{\makebox[1.9em][r]{#1}\makebox[2.1em][l]{\hspace{0.12em}\raisebox{-0.22em}[0pt][0pt]{\fontsize{6}{6}\selectfont\color{#2}#3}}}
\newcommand{\rwup}[2]{\rwd{#1}{ForestGreen}{+#2}}
\newcommand{\rwdn}[2]{\rwd{#1}{BrickRed}{\textminus#2}}
\begin{tabular*}{\columnwidth}{@{}l@{\extracolsep{\fill}}wc{0.235\columnwidth}@{\extracolsep{0pt}}wc{0.245\columnwidth}wc{0.245\columnwidth}@{}}
\toprule[0.9pt]
\textbf{Backbone} & No recovery & ViRA-ZS & \cellcolor{blue!6}\textbf{ViRA} \\
\midrule
Qwen3.5-9B  & 78.7 & \rwdn{78.1}{0.6} & \cellcolor{blue!6}\rwup{\textbf{81.2}}{2.5} \\
Qwen3-VL-8B & 74.4 & \rwdn{72.6}{1.8} & \cellcolor{blue!6}\rwup{\textbf{75.6}}{1.2} \\
LLaVA-OV2-8B & 68.6 & \rwup{\textbf{69.4}}{0.8} & \cellcolor{blue!6}--- \\
\midrule
\textbf{Backbone} & No recovery & Robust-R1 (SFT) & Robust-R1 (RL) \\
\midrule
Qwen2.5-VL-3B & \textbf{65.1} & \rwdn{63.1}{2.0} & \rwdn{62.7}{2.4} \\
\bottomrule[0.9pt]
\end{tabular*}
\end{table}

%% file: tables/clean_table.tex
\begin{table}[t]
\centering
\caption{
\textbf{Performance comparison on clean videos.}
We report accuracy (\%) of Qwen3.5-9B, ViRA-ZS, and ViRA on uncorrupted inputs. Best results are shown in \textbf{bold}. Both ViRA-ZS and ViRA outperform the base model across all five evaluation subsets, demonstrating that agentic recovery introduces no clean-data performance trade-off.}
\vspace{-3pt}
\label{tab:clean}
\setlength{\tabcolsep}{8pt}
\resizebox{\columnwidth}{!}{%
\begin{tabular}{lccccc}
\toprule
\multirow{2}{*}{\textbf{Method}}
& \multicolumn{3}{c}{\textbf{Video-MME}}
& \multirow{2}{*}{\textbf{STAR}}
& \multirow{2}{*}{\textbf{NExT-QA}} \\
\cmidrule(lr){2-4}
& \textbf{Short}
& \textbf{Medium}
& \textbf{Long}
& \\
\midrule
Qwen3.5-9B & 71.8 & 58.6 & 53.0 & 69.0 & 81.8 \\
ViRA-ZS    & 74.3 & 60.2 & \textbf{55.0} & 69.8 & 82.2 \\
ViRA       & \textbf{74.4} & \textbf{60.7} & \textbf{55.0} & \textbf{71.2} & \textbf{83.4} \\
\bottomrule
\end{tabular}%
}
\vspace{-10pt}
\end{table}

%% file: sec/5_conclusion.tex
\section{Conclusion}
\label{sec:conclusion}

ViRA is a video recovery agent for robust video understanding under visual corruption. Instead of fixed preprocessing, ViRA integrates recovery into the reasoning loop through a diagnose--restore--reason paradigm: it determines whether recovery is needed, selects restoration tools and their temporal intervals, and reasons over the recovered evidence. This query-conditioned design focuses computation on task-relevant degraded regions while avoiding recovery when the original evidence is sufficient.
Across multiple video understanding benchmarks, corruption regimes, and MLLM backbones, ViRA consistently improves performance over direct inference on degraded videos while largely preserving clean-input performance. These results highlight the value of reasoning about when and how to recover visual evidence.